\UseRawInputEncoding
\documentclass[lettersize,journal]{IEEEtran}
\usepackage{amsmath,amsfonts}
\usepackage{algorithmic}
\usepackage{algorithm}
\usepackage{array}
\usepackage[caption=false,font=normalsize,labelfont=sf,textfont=sf]{subfig}
\usepackage{textcomp}
\usepackage{stfloats}
\usepackage{url}
\usepackage{verbatim}
\usepackage{graphicx}
\usepackage{cite}
\usepackage{amsthm, amssymb}
\theoremstyle{definition}
\newtheorem{definition}{Definition}[section]
\usepackage{booktabs}
\usepackage{longtable}
\usepackage{multirow}
\usepackage{makecell}
\usepackage{xcolor}
\begin{document}

\title{KoopAGRU: A Koopman-based Anomaly Detection in Time-Series using Gated Recurrent Units}

\author{
Issam Ait Yahia, Ismail Berrada 
% ~\IEEEmembership{Staff,~IEEE,}
        % <-this % stops a space
\thanks{Issam Ait Yahia, College of Computing,
       Mohammed VI Polytechnic University,
       Ben Guerir, Morocco,  issam.aityahia@um6p.ma.
       
       Ismail Berrada, College of Computing,
        Mohammed VI Polytechnic University,
        Ben Guerir, Morocco, ismail.berrada@um6p.ma.}}% <-this % stops a space

% \markboth{This work has been submitted to the IEEE TPAMI for possible publication. Copyright may be transferred without notice, after which this version will no longer be accessible.}%
\markboth{Submitted to IEEE TPAMI for possible publication. Copyright may be transferred.}
{Shell \MakeLowercase{\textit{et al.}}: A Sample Article Using IEEEtran.cls for IEEE Journals}


\maketitle

\begin{abstract}
Anomaly detection in real-world time-series data is a challenging task due to the complex and nonlinear temporal dynamics involved. This paper introduces KoopAGRU, a new deep learning model designed to tackle this problem by combining  Fast Fourier Transform (FFT), Deep Dynamic Mode Decomposition (DeepDMD), and Koopman theory. FFT allows KoopAGRU to decompose temporal data into time-variant and time-invariant components providing precise modeling of complex patterns. To better control these two components, KoopAGRU utilizes Gate Recurrent Unit (GRU) encoders to learn Koopman observables, enhancing the detection capability across multiple temporal scales. KoopAGRU is trained in a single process and offers fast inference times. Extensive tests on various benchmark datasets show that KoopAGRU outperforms other leading methods, achieving a new average F1-score of 90.88\% on the well-known anomalies detection task of times series datasets, and proves to be efficient and reliable in detecting anomalies in real-world scenarios.
\end{abstract}


\section{Introduction}

\IEEEPARstart{I}{n} real-world cyber-physical systems such as industrial equipment, space probes, and smart factories, a multitude of sensors operate continuously, generating substantial amounts of sequential measurements. To effectively monitor these systems' real-time conditions, it is crucial to detect anomalies that may indicate potential risks or impending financial losses. This process, known as time-series anomaly detection, involves identifying abnormal system states within each time step of the collected data.

A robust time series anomaly detection method requires accurate modeling of the complex and nonlinear temporal dynamics of normal system behaviors while generalizing to unforeseen anomalies. This task is challenging due to the complex interactions among variables and highly nonlinear temporal dependencies in real-world data. Additionally, the rarity of anomalies makes their identification and labeling time-consuming and costly. Consequently, time-series anomaly detection is typically approached as an unsupervised learning problem.

To address the previous challenges, several advanced techniques and theories introduced in the field of dynamic systems could offer valuable insights and inspiration. In fact, Koopman operator theory \cite{mezic2020koopmanoperatorgeometrylearning} simplifies the study of nonlinear dynamical systems by transforming them into higher-dimensional spaces, thus facilitating the identification of complex behaviors indicative of anomalies \cite{Brunton2016}. Deep Dynamic Mode Decomposition (DeepDMD) \cite{deepdmd} enhances this approach by employing deep learning to directly identify Koopman operators and observable functions from data \cite{Takeishi2017}. The incorporation of Gated Recurrent Unit (GRU) layers captures long-term dependencies and temporal relationships, which are crucial for accurate anomaly detection \cite{Chung2014}. In particular, the use of a GRU encoder enhances the model's ability to learn data representations effectively, capturing the temporal dynamics essential for identifying anomalies \cite{Harrou2024}. On the other hand, the Fast Fourier Transform (FFT) \cite{5217220} decomposes time-series data into frequency components, distinguishing between time-variant and time-invariant elements. Recent research on the Non-Stationary Transformer \cite{nsformer} highlights the FFT's effectiveness for anomaly detection in non-stationary data by revealing irregular patterns and deviations that are otherwise hidden in the time domain.

This paper introduces a novel state-of-the-art anomaly detection approach that combines FFT, DeepDMD, and GRU encoders. The proposed model, namely KoopAGRU,  leverages the strengths of each component to tackle the complexities of time-series data. Precisely, FFT decomposes the data into its frequency components, DeepDMD captures the underlying dynamics of the system, and the GRU encoder learns Koopman observables within DeepDMD to handle sequential dependencies and temporal patterns effectively, isolating anomalies across various frequency domains and temporal patterns. To validate our approach, extensive experiments were conducted on various benchmark datasets, including the Server Machine Dataset (SMD) \cite{smd}, Mars Science Laboratory (MSL) \cite{mslsmap}, Soil Moisture Active Passive (SMAP) \cite{mslsmap}, Secure Water Treatment (SWaT) \cite{swat}, and Public Safety Monitoring (PSM) \cite{psm}. These datasets cover a wide range of applications and anomaly types, offering a thorough evaluation of the model's performance. KoopAGRU significantly outperforms other leading methods, achieving a new average F1-score of 90.37\% on these datasets. It also exhibits high robustness and effectiveness in anomaly detection tasks, handling the complex and nonlinear nature of real-world time-series data with fewer parameters, reduced resource usage, and faster inference times.
\IEEEpubidadjcol
Thus, the paper's contributions are summarized as follows:
\begin{itemize}
    \item Introduction of a novel state-of-the-art forecasting model for time-series anomaly detection. KoopAGRU achieves an F1-score of 90.88 in anomaly detection tasks, demonstrating its effectiveness in identifying unusual patterns or anomalies within time series data.
    \item Introduction of a new hyperparameter $\beta$, to control the influence of the time-invariant component, allowing for finer tuning and better adaptation to different types of time series data.
    \item Construction of observables directly from measurements to simplify the model by eliminating the need for inverse mapping, and thus reducing computational complexity.
    \item The design of the Koopman operator with a fixed matrix size allowing the building of small models while maintaining high performance, scalability, and deployability.
\end{itemize}

%%%%%%%%%%%%%%%%%%%%%%%%%%%%%%%%%%%%%% Related Work %%%%%%%%%%%%%%%%%%%%%%%%%%%%%%%%%%%%
\section{Related Work}
The analysis and forecasting of time series data have garnered considerable attention across various disciplines, leading to substantial progress in anomaly detection, prediction, and signal representation. This section provides a comprehensive overview of existing methodologies in unsupervised time series anomaly detection, the application of Koopman theory for nonlinear system modeling and prediction, and the utilization of the Fast Fourier Transform (FFT) in signal processing and representation learning.  Notably, existing Koopman-based models have primarily been utilized for long-term forecasting tasks. The present work aims to address this limitation by extending the applicability of Koopman theory to unsupervised anomaly detection and short-term forecasting, thus filling a critical gap in the current literature.
\subsection{Unsupervised Time Series Anomaly Detection}

While unsupervised time series anomaly detection is a critical and widely studied problem, proposed approaches can be broadly categorized into four categories:
\begin{enumerate}
    \item \textit{Density-estimation methods} include techniques that span local density (e.g. Local Outlier Factor \cite{inproceedings1}), connectivity (e.g. Connectivity Outlier Factor \cite{10.1007/3-540-47887-6_53}), and Gaussian Mixture Models (e.g. DAGMM \cite{zong2018deep} and MPPCACD \cite{7859276}).

\item \textit{Clustering-based methods} typically rely on specific criteria (e.g. distance from data points to the cluster center) to define the anomaly score, and normal data to form compact clusters (e.g. SVDD \cite{article} and Deep SVDD \cite{inproceedings}).  THOC \cite{NEURIPS2020_97e401a0} integrates multi-scale temporal features through hierarchical clustering. ITAD \cite{shin2020integrative} performs clustering on decomposed tensors.

\item \textit{Reconstruction-based models}  on the other hand, rely on evaluating the reconstruction errors. \cite{park2017multimodalanomalydetectorrobotassisted} use an LSTM-VAE model for reconstruction, while \cite{Su2019RobustAD} extend it by incorporating normalizing flows. The detection is then based on reconstruction probabilities. \cite{10.1145/3447548.3467075} employ a hierarchical VAE to capture inter/intra-dependencies among multiple time series. Finally, \cite{articleGAN} propose the use of GANs \cite{goodfellow2014generativeadversarialnetworks} for reconstruction-based anomaly detection. Similarly, \cite{fedanids} leverage autoencoders in a federated learning framework for NIDS, while \cite{domainadapt} and \cite{nfnids} explore domain adaptation and normalizing flows, respectively, for effective NIDS reconstruction and detection.

\item  Autoregression (AR) models have been widely utilized in anomaly detection, leveraging their ability to model temporal dependencies in time series data. \cite{box2015timeseriesanalysis} and \cite{hamilton1994timeseriesanalysis} laid the groundwork for AR models, which have since been adapted for detecting anomalies by modeling expected behavior and identifying deviations. Recent works, such as \cite{shih2019multivariatetimeseriesforecasting} and \cite{lim2021arnettemporal}, have enhanced traditional AR models with mechanisms like temporal attention and deep learning, improving their effectiveness in complex anomaly detection tasks.
\end{enumerate}

\subsection{Koopman Theory in Nonlinear Systems Analysis and Prediction}

Recently, Koopman theory has been actively applied to analyzing modern dynamical systems \cite{doi:10.1073/pnas.17.5.315,strogatz2018nonlinear}. The Koopman Operator Theory (KOT) maps nonlinear systems as globally linear systems in a space of observable functions. Although the Koopman operator is infinite-dimensional, it can be approximated using finite-dimensional methods such as Dynamic Mode Decomposition (DMD) \cite{brunton2016extracting,rowley2009spectral,tu2013dynamic}, extended DMD \cite{williams2015data}, robust DMD \cite{shin2020integrative}, and deep DMD \cite{takeishi2017learning, li2017extended, lusch2018deep, yeung2019learning, 9738598}.

Koopman-based techniques are widely used in fluid mechanics \cite{rowley2009spectral}, synthetic biology \cite{yeung2019learning, eisenhower2010decomposing}, and energy systems \cite{sinha2020data, susuki2016applied}. Applications in power systems include nonlinear modal analysis for coherency detection \cite{susuki2016applied}, attack identification \cite{nandanoori2020model}, and anomaly diagnosis \cite{surana2020koopman}. Some studies focus on time-series prediction \cite{susuki2016applied}. \cite{nandanoori2020model} employ DMD for transient prediction in power systems, while other works combined KOT with deep learning tools for improved prediction models \cite{netto2021analytical}.

Recent advancements integrate deep learning into the Koopman theory enabling data-driven approaches such as:
\begin{itemize}
    \item Koopman Autoencoders \cite{erichson2019physics, Lusch_2018, takeishi2018learningkoopmaninvariantsubspaces} learns measurement functions and operators simultaneously. 
    \item Sparse identification of nonlinear dynamics for model predictive control in the low-data limit  \cite{Kaiser_2018} introduces a backward procedure to improve operator consistency and stability. 
    \item Koopman spectral analysis \cite{lange2020fourierkoopmanspectralmethods} facilitates sequence prediction by disentangling dominant factors. 
    \item Koopman Neural Forecaster (KNF) \cite{wang2023koopmanneuralforecastertime} learns Koopman operators and attention maps for time series forecasting with varying distributions.

    \item Modular Koopman Predictors \cite{koopa2022} address time-variant and time-invariant components with hierarchically learned operators, innovating by removing reconstruction loss in the Koopman Autoencoder for fully predictive training. This approach allows the Koopman matrix to evolve over time for each time series, adapting to changing distributions. 

    \item Traffic Patterns \cite{avila2020} utilized KMD to analyze and forecast highway traffic dynamics, reconstructing observed data and identifying both known and novel spatiotemporal patterns for improved network condition predictions.
    
    \item Disease Predictions \cite{brunton2017} demonstrated the use of Koopman operator theory in modeling and forecasting the dynamics of epidemiological systems, while \cite{mezic2024} expanded this work to further refine disease outbreak predictions using Koopman-based approaches.

    \item Sea Ice Prediction \cite{hogg2020} applied Koopman Mode Decomposition (KMD) to satellite data of sea ice concentration, uncovering spatial modes and providing accurate geographic predictions of sea ice concentration up to four years into the future.

\end{itemize}
\subsection{Fast Fourier Transform (FFT)}
The Fast Fourier Transform (FFT) is an essential algorithm in signal processing, used to decompose time-domain signals into their frequency components efficiently. This capability is critical for signal analysis, noise reduction, and feature extraction in fields such as telecommunications and medical imaging. FFT has recently gained prominence in machine learning, particularly in representation and deep learning. For example, FFT transforms time-series data into the frequency domain, revealing patterns that improve model performance in tasks like EEG analysis \cite{kiranyaz2021self}. Additionally, FFT accelerates convolution operations in Convolutional Neural Networks (CNNs), reducing computational complexity \cite{mathieu2014fast}. Hybrid models combining FFT with neural networks have also emerged, enhancing the ability to learn from frequency-domain data \cite{denoise2018}.

%%%%%%%%%%%%%%%%%%%%%%%%%%%%%%%%%%%%%% Methodology %%%%%%%%%%%%%%%%%%%%%%%%%%%%%%%%%%%%
\section{Methodology}
This section introduces KoopAGRU, a novel state-of-the-art deep learning model for time series anomaly detection. Our methodology integrates Koopman operator theory with both Deep Dynamic Mode Decomposition (DeepDMD) and Fourier analysis to effectively model and represent time series data. Koopman theory provides a strong framework for linearizing nonlinear dynamical systems by transforming them into a higher-dimensional space, facilitating the analysis and prediction of complex behaviors. For additional information on Koopman theory, refer to Appendix \ref{appendix:koopman}. DeepDMD extends this framework by using deep learning techniques to identify Koopman operators and the associated observable functions from data. Further details on DeepDMD can also be found in Appendix \ref{appendix:DeepDMD}.

\begin{figure*}[h]
\centering
  \includegraphics[width=0.9\textwidth]{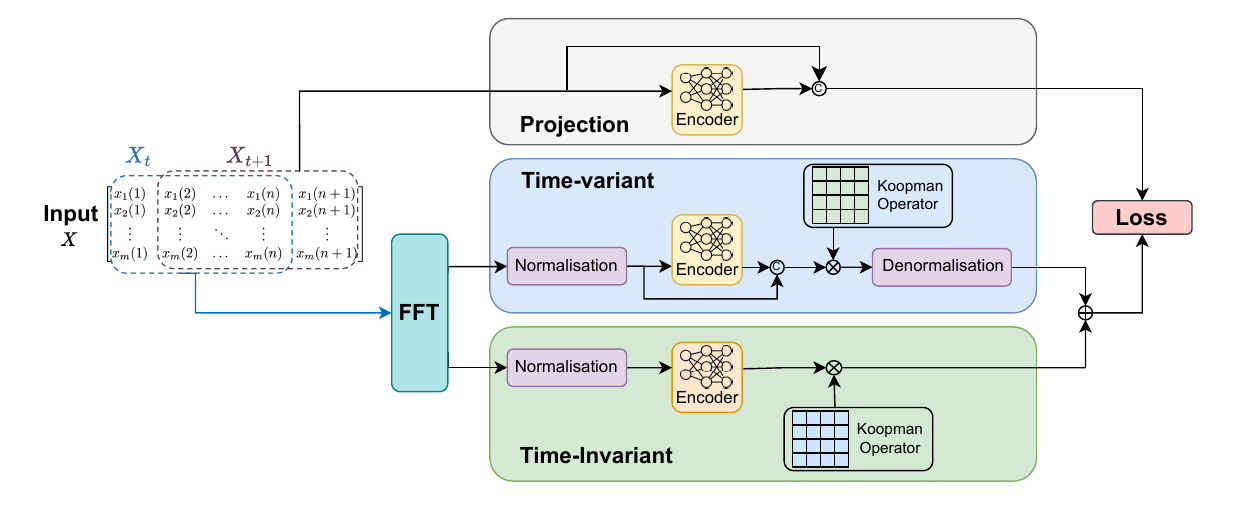}
  \caption{KoopAGRU: Overall Model Architecture.}
  \label{fig:diag}
\end{figure*}

\subsection{KoopAGRU Overview}
Figure\,\ref{fig:diag} illustrates a global overview of KoopAGRU. To process time-series data, KoopAGRU combines several primary components:

\begin{itemize}
    \item Input: KoopAGRU processes a sequence of measurements \(X\), splitting it into \(X_t\) (the present  measurements) and \(X_{t+1}\) (the future measurements). By shifting the data, the model captures temporal relationships within the sequence. Each part of the sequence is then directed to specific components for further processing.

    \item Fast Fourier Transform (FFT):  Fourier analysis is used to identify dominant and varying frequency spectra, separating time-invariant and time-variant components of a sequence of measurements (\(X_t\) and \(X_{t+1}\)). Dominant frequencies represent globally shared dynamics while varying frequencies capture localized variations.

    \item Time-Variant: This component captures the dynamic aspects of the time series that fluctuate over time. By isolating the time-variant fragments using Fourier analysis, the model enhances its ability to encode and predict the evolving dynamics of the system.

    \item Time-Invariant: This component represents the stable, unchanging aspects of the time series. By focusing on the time-invariant parts, the model retains the fundamental characteristics of the data that remain constant over time, providing a stable foundation for analysis.

    \item Projection: This component is used during training to map \(X_{t+1}\), the future measurements, into a function space. This approach allows the model to eliminate the need for a decoder, simplifying the process while still enabling effective analysis and prediction.

    \item Loss: The loss function measures the difference between the predicted outputs and the actual targets. It plays a crucial role in guiding model training, helping to improve accuracy and refine the model's anomaly detection capabilities.
\end{itemize}

\subsection{Input}
The input matrix \( X \) can be represented as a collection of column vectors \( \mathbf{x}(1), \mathbf{x}(2), \dots, \mathbf{x}(n+1) \), where each column vector \( \mathbf{x}(i) \) is defined as:
\begin{equation}
\mathbf{x}(i) = \begin{pmatrix} x_1(i) \\ x_2(i) \\ \vdots \\ x_m(i) \end{pmatrix}
\end{equation}
\\
Here, \( \mathbf{x}(i) \) denotes the state of all time series at a specific time point \( i \), effectively capturing both individual time series values and the overall state across multiple time series.
KoopAGRU processes these measurements by dividing them into \( \mathbf{X}_t \) (the past) and \( \mathbf{X}_{t+1} \) (the future). Specifically:
\begin{equation}
\mathbf{X}_t = \begin{pmatrix} \mathbf{x}(1) & \mathbf{x}(2) & \dots & \mathbf{x}(n) \end{pmatrix}
\end{equation}
\begin{equation}
\mathbf{X}_{t+1} = \begin{pmatrix} \mathbf{x}(2) & \mathbf{x}(3) & \dots & \mathbf{x}(n+1) \end{pmatrix}
\end{equation}
\\
By shifting the data, the model captures temporal relationships within the sequence.  \( \mathbf{X}_t \) and \( \mathbf{X}_{t+1} \) are then directed to a specific component of KoopAGRU for further analysis, enabling the model to process and understand temporal dynamics effectively.
\subsection{Fast Fourier Transform (FFT)}

Fourier analysis is employed to identify globally shared and localized frequency spectrums across different periods, separating the components of the series. FFT is precomputed for each lookback window in the training set. The averaged amplitude of each spectrum \( S = \{0, 1, \ldots, \lfloor T/2 \rfloor\} \) is calculated and ranked by amplitude. The top \( \alpha \) percent are selected as the subset \( G_\alpha \subset S \), representing dominant spectrums common to all lookback windows and indicative of time-invariant dynamics. The remaining spectrums pertain to varying windows over different periods, dividing the spectrums \( S \) into \( G_\alpha \) and its complement \( \bar{G}_\alpha \). 
\\
\\
During training and inference, the FFT component of KoopAGRU separates the input \( X_t \) as follows:

\begin{equation}\label{eq1}
\begin{aligned}
X_{\text{inv}} &= \mathcal{F}^{-1} \left( \text{F}_f\left( G_\alpha, \mathcal{F}(X_t) \right) \right) \\
X_{\text{var}} &= \mathcal{F}^{-1} \left( \text{F}_f\left( \bar{G}_\alpha, \mathcal{F}(X_t) \right) \right) \\
               &= X_t - X_{\text{inv}}
\end{aligned}
\end{equation} \\
where \( \mathcal{F} \) denotes FFT, \( \mathcal{F}^{-1} \) is its inverse, and \(\text{F}_f(\cdot)\) is the filter that permits only the frequency spectrums specified by the given set. This method is adopted from the Koopa paper \cite{koopa2022}.

\subsection{Time-Variant}

\begin{figure}[h]
\centering
  \includegraphics[width=1.05\columnwidth]{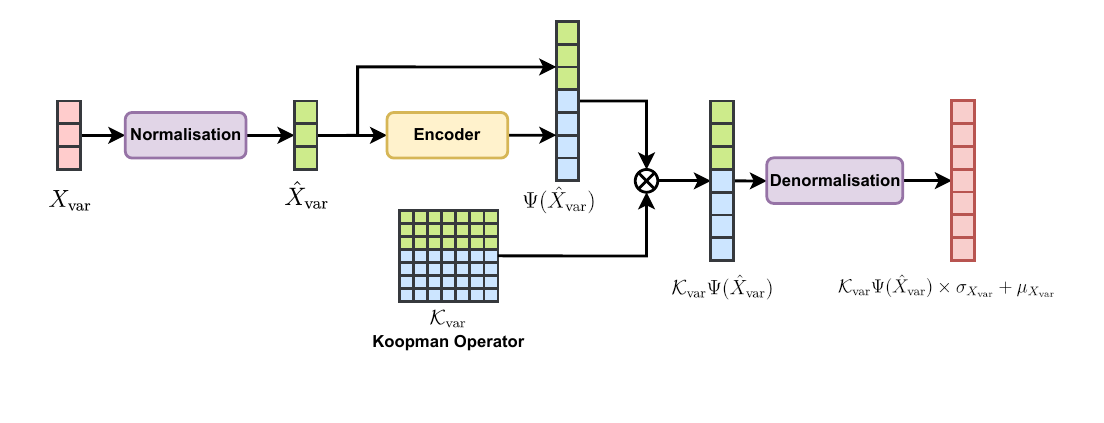}
  \caption{Time-Variant Component Architecture.}
  \label{fig:diag_time_var}
\end{figure}
The time-variant component of KoopAGRU captures the dynamic aspects of the data that fluctuate over time. Using the Fourier filter, these components are isolated, enhancing the model's ability to encode the temporal dynamics of the series. The time-variant component is denoted by \( X_{\text{var}} \):

\begin{equation}\label{eq3}
X_{\text{var}} = \mathcal{F}^{-1} \left( \text{F}_f\left( \bar{G}_\alpha, \mathcal{F}(X_t) \right) \right)   
\end{equation}
\\
The time-variant component spans four main blocks: Normalisation, Encoder, Koopman Operator, and Denormalisation (Figure \ref{fig:diag_time_var}). 
\subsubsection{Normalisation}
Following the approach used in \cite{nsformer}, we apply normalisation to the time-variant component to ensure stability and consistency during the encoding process. This method eliminates biases caused by varying scales in the data by centering the values and standardizing their scale, leading to robust and reliable representations of time-variant dynamics. The normalisation is performed as:
\begin{equation}\label{eq4}
\hat{X}_{\text{var}} = \frac{X_{\text{var}} - \mu_{\scriptscriptstyle X_{\text{var}}}}{\sigma_{\scriptscriptstyle X_{\text{var}}}}
\end{equation} \\
where \(\mu_{\scriptscriptstyle X_{\text{var}}}\) is the mean and \(\sigma_{\scriptscriptstyle X_{\text{var}}}\) is the standard deviation of the time-variant component.

\subsubsection{Encoder}
The encoder for the time-variant component transforms the normalized data into a higher-dimensional space suitable for further processing. 
\\
\\
Let \( X_t \in \mathbb{R}^n \) represent the \( n \)-dimensional measurements available at any given time \( t \). We define \(\psi : \mathbb{R}^n \to \mathbb{R}^q \) as a \( q \)-dimensional vector-valued observable function of the measurements \( X_t \), described by:
\begin{equation}\label{eq5}
\psi(X_t) := [\psi_1(X_t), \psi_2(X_t), \ldots, \psi_q(X_t)]^\top
\end{equation} \\
where each \(\psi_i\) is a scalar-valued observable function of the measurements \( X_t \). The encoder's task is then to map the time-variant input data into this observable space, capturing the temporal dynamics in a way that is conducive to linear analysis by the Koopman operator.

\subsubsection{Koopman Operator}
The Koopman operator denoted as \( \mathcal{K}_{\text{var}} \), is a linear operator that governs the evolution of the encoded observables over time. The objective is to learn this linear operator such that it can predict the future states of the observables. The transformed time-variant component using the Koopman operator \( \mathcal{K}_{\text{var}} \) is represented as:
\begin{equation}
\mathcal{K}_{\text{var}} \Psi(\hat{X}_{\text{var}})
\end{equation} \\
where \( \Psi(X_t) \) includes both the original measurements and the encoded observables:
\begin{equation}
\Psi(X_t) = \begin{bmatrix} X_t \\ \psi(X_t) \end{bmatrix}
\end{equation} \\
Since the measurements are directly accessible, inverse mapping is unnecessary for measurement-inclusive observable functions. Due to the simplicity of this approach, we adopted it for constructing measurement-inclusive observables. This approach is similar to the concept of state-inclusive observables discussed in references \cite{johnson2018logistic, yeung2019learning}.
By applying the Koopman operator to the encoded representation, we can leverage the power of linear dynamics to model the complex temporal evolution of the system.

\subsubsection{Denormalisation}

Finally, denormalisation is applied to revert the transformed data back to its original scale:
\begin{equation}
\mathcal{K}_{\text{var}} \Psi(\hat{X}_{\text{var}}) \times \sigma_{\scriptscriptstyle X_{\text{var}}} + \mu_{\scriptscriptstyle X_{\text{var}}} 
\end{equation}

\subsection{Time-Invariant}
\begin{figure}[h]
\centering
  \includegraphics[width=\columnwidth]{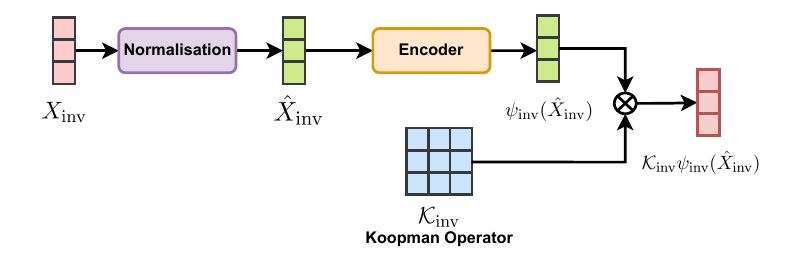}
  \caption{Time-Invariant Component Architecture.}
  \label{fig:diag_time_invar}
\end{figure}
% \textcolor{blue}{Add the image and list the components}
The time-invariant component of KoopAGRU represents the stable aspects of the time series that remain constant over time. Isolating these components with the Fourier filter ensures the stability of the encoded representation. The time-invariant component is denoted by \( X_{\text{inv}} \):

\begin{equation}
X_{\text{inv}} = \mathcal{F}^{-1} \left( \text{F}_f\left( G_\alpha, \mathcal{F}(X_t) \right) \right) 
\end{equation} \\
The time-invariant component spans three main blocks: Normalisation, Encoder, and Koopman
Operator (Figure \ref{fig:diag_time_invar}). 

\subsubsection{Normalisation}

Normalisation is applied to the time-invariant component to maintain stability and consistency in the encoding process:
\begin{equation}
\hat{X}_{\text{inv}} = \frac{X_{\text{inv}} - \mu_{\scriptscriptstyle X_{\text{inv}}}}{\sigma_{\scriptscriptstyle X_{\text{inv}}}}
\end{equation}

\subsubsection{Encoder}

The encoder network processes the time-invariant components and extracts meaningful features. Given the inherent simplicity of the time-invariant component compared to the time-variant component, only an encoder is utilized for this part of the model. The normalized time-invariant input \(\hat{X}_{\text{inv}}\) is fed into the encoder to generate the encoded representation:
\begin{equation}
\psi_{\text{inv}}(\hat{X}_{\text{inv}})
\end{equation}

\subsubsection{Koopman Operator}
The encoded representation is then transformed using the Koopman operator \(\mathcal{K}_{\text{inv}}\), which captures the invariant dynamics:

\begin{equation}
\mathcal{K}_{\text{inv}}  \psi_{\text{inv}}(\hat{X}_{\text{inv}})
\end{equation}
To match the dimension of the variant component, we add zeros to the time-invariant component if necessary. 
\begin{equation}
\Psi(\hat{X}_{\text{inv}}) = \begin{bmatrix} \mathcal{K}_{\text{inv}}  \psi_{\text{inv}}(\hat{X}_{\text{inv}}) \\ 0 \end{bmatrix}
\end{equation}
It is important to note that denormalisation is not required after applying the Koopman operator, as the ablation studies presented in Table \ref{tab:norm_denorm} demonstrate that the best results are obtained without the denormalisation step.
\subsection{Projection}
The projection operation constitutes a key element of the model, applied solely during the training phase. This operation projects the future time-series component into an observable space, where raw measurements are combined with derived observables that capture essential system dynamics. This structure is represented as follows:

\begin{equation}
\Phi_{X_{t+1}} = \Psi(X_{t+1}) = \begin{bmatrix} X_{t+1} \\ \psi(X_{t+1}) \end{bmatrix}
\end{equation} 
This combined form, termed a "measurement-inclusive observable vector," 
integrates direct measurements and corresponding observables into a cohesive representation.

The primary purpose of limiting this projection to the training phase is to simplify the model architecture by eliminating the need for a decoder. By mapping the loss function into this observable space, the model is able to learn the dynamics effectively without necessitating a decoding step, thus maintaining architectural simplicity. Notably, the encoder employed here is the same as that used for the time-variant components, leveraging shared weights to capture consistent dynamic features across these dimensions. Consequently, during the testing phase, this projection is no longer required. With direct access to information in the measurement-inclusive observables, the model can predict future values effectively, relying solely on the learned dynamics without reintroducing either the projection or decoder components. For a detailed exposition of this approach, please refer to the appendix \ref{appendix:DeepDMD}.

\subsection{Loss}

The loss function employed in KoopAGRU comprises several components to ensure effective model training. The comprehensive loss function is articulated as follows:

\begin{equation}
\mathcal{L} = \mathcal{L}_{\text{koopman}} + \lambda \mathcal{L}_{\text{reg}}
\label{eq_lambda}
\end{equation}
where,
\begin{equation}
\mathcal{L}_{\text{koopman}} = \| \Phi_{X_{t+1}} - \Phi_{X_{t}} \|_{\scriptscriptstyle\mathcal{F}}
\end{equation}
and
\begin{equation}
\mathcal{L}_{\text{reg}} = \| \mathcal{K}_{\text{var}} \|_{\scriptscriptstyle\mathcal{F}} + \| \mathcal{K}_{\text{inv}} \|_{\scriptscriptstyle\mathcal{F}} \\
\end{equation}
\( \mathcal{L}_{\text{koopman}} \) represents the Frobenius norm of the difference between the encoded representations \(\Phi_{X_{t+1}}\) (future state) and \(\Phi_{X_{t}}\) (predicted state). The Frobenius norm is efficient in this context because it measures the element-wise differences between matrices, offering robustness to the error in the Koopman framework. This formulation ensures that the Koopman operator accurately models the system dynamics by minimizing the prediction error. The term \(\Phi_{X_{t}}\) is defined as follows:

\begin{equation}
\Phi_{X_{t}} = \beta \times \Psi(\hat{X}_{\text{inv}}) + \mathcal{K}_{\text{var}} \Psi(\hat{X}_{\text{var}}) \times \sigma_{X_{\text{var}}} + \mu_{X_{\text{var}}}
\label{eq_beta}
\end{equation}
It represents the sum of the time-invariant and time-variant components, normalized and transformed by their respective Koopman operators. Specifically, this captures both the invariant and dynamic aspects of the system. The parameter \(\beta\) is used to control the influence of the time-invariant component in the model.

The regularization term \( \mathcal{L}_{\text{reg}} \) penalizes the Frobenius norms of the Koopman operator matrices \( \mathcal{K}_{\text{var}} \) and \( \mathcal{K}_{\text{inv}} \), which correspond to the variant and invariant components, respectively. The parameter \(\lambda\) is a regularization coefficient that controls the weight of this penalty term. A suitable choice of \(\lambda\) is essential as it balances the trade-off between fitting the training data and maintaining model simplicity. Setting \(\lambda\) too high can overly constrain the model, leading to underfitting, while setting it too low can result in overfitting by allowing excessive model complexity. Thus, \(\lambda\) is a crucial hyperparameter that must be carefully tuned to ensure the Koopman operator matrices are both effective and generalizable.

%%%%%%%%%%%%%%%%%%%%%%%%%%%%%%%%%%%%%% Experiments %%%%%%%%%%%%%%%%%%%%%%%%%%%%%%%%%%%%

\section{Experiments}
\subsection{DataSets}
The evaluation of KoopAGRU was conducted using the five datasets listed in Table \ref{tab:datasets}:

\begin{itemize}
    \item \textbf{SMD} (\textbf{S}erver \textbf{M}achine \textbf{D}ataset). This dataset, collected from server machines, includes 38 dimensions and comprises 566,724 training samples, 141,681 validation samples, and 708,420 test samples.
    \item \textbf{PSM} (\textbf{P}ooled \textbf{S}erver \textbf{M}etric). This dataset contains 25 dimensions and consists of 105,984 training samples, 26,497 validation samples, and 87,841 test samples.
    \item \textbf{MSL} (\textbf{M}ars \textbf{S}cience \textbf{L}aboratory). This dataset, derived from space exploration operations, comprises 55 dimensions, with 44,653 training samples, 11,664 validation samples, and 73,729 test samples.
    \item \textbf{SMAP} (\textbf{S}oil \textbf{M}oisture \textbf{A}ctive \textbf{P}assive). This dataset includes 25 dimensions and contains 108,146 training samples, 27,037 validation samples, and 427,617 test samples.
    \item \textbf{SWaT} (\textbf{S}ecure \textbf{Wa}ter \textbf{T}reatment). This dataset, related to water treatment operations, comprises 51 dimensions and includes 396,000 training samples, 99,000 validation samples, and 449,919 test samples.
\end{itemize}
\begin{table}[h!]
\caption{Evaluation datasets.}
\centering
\footnotesize 
\begin{tabular}{c|c|c|c|c|c}
\toprule
\hline
Dataset & Domain & Dimension & \#Training & \#Validation & \#Tests \\
\hline
SMD & Server & 38 & 566724 & 141681 & 708420 \\
PSM & Server & 25 & 105984 & 26497 & 87841 \\
MSL & Space & 55 & 44653 & 11664 & 73729  \\
SMAP & Space & 25 & 108146 & 27037 & 427617 \\
SWaT & Water & 51 & 396000 & 99000 & 449919  \\
\hline
\bottomrule
\end{tabular}

\label{tab:datasets}
\end{table}

\subsection{Anomaly Detection Approach}

Anomaly detection in time series data is critical for industrial monitoring and maintenance, as anomalies often indicate potential faults or irregularities in system operations. Given the large volume of such data, manual labeling is not feasible, making unsupervised detection methods indispensable. In this context, a forecasting model is adapted for anomaly detection by utilizing its next-step prediction capabilities. This method is inspired by and adapted from the approach proposed in the \cite{IDRISSI2023121000}.

At each time step \( t \), the model predicts the next value in the time series, \( \hat{X}_{t+1} \), based on observed data up to \( t \). The discrepancy between the predicted value \( \hat{X}_{t+1} \) and the actual observed value \( X_{t+1} \) is calculated as the prediction error \( e_t \), defined as follows:

\begin{equation}
    e_t = \| X_{t+1} - \hat{X}_{t+1} \|
\end{equation}

Here, \( \| \cdot \| \) represents the chosen norm (in this case, the \( L_2 \) norm), which quantifies the magnitude of the deviation. This prediction error is the primary metric used for detecting anomalies.

During the validation phase, a threshold \( \delta \) is determined to differentiate normal behavior from anomalies. To achieve flexibility and robustness, \( \delta \) is computed using a percentile-based method. This method establishes \( \delta \) as the value corresponding to the \( r \)-th percentile of the prediction error distribution \( \{ e_t \} \) from the validation set:

\begin{equation}
    \delta = P_r(e)
    \label{eq10}
\end{equation}

where \( P_r(e) \) denotes the \( r \)-th percentile of the prediction errors. The value of \( r \) is dataset-specific, chosen to calibrate the proportion of validation data labeled as anomalies. This percentile-based thresholding ensures that \( \delta \) adapts to the variability and scale of prediction errors in each dataset. During testing, any time step \( t \) is flagged as an anomaly if the corresponding prediction error satisfies the condition:

\begin{equation}
    e_t > \delta
\end{equation}

By dynamically adjusting \( \delta \) based on the validation data, this method remains robust across datasets with varying characteristics. Additionally, the flexibility to control \( r \) allows for fine-tuning of anomaly sensitivity, ensuring that significant deviations indicative of faults are effectively detected. This approach enhances the scalability and reliability of anomaly detection in large-scale industrial systems.
\subsection{Implementation details}
\paragraph{\textbf{Preprocessing}.} The preprocessing pipeline follows the methodology introduced in \cite{anomaly2021}, utilizing a non-overlapping sliding window to segment datasets into sub-series. The sliding window size is fixed at 100 for all datasets. Anomalies are identified by comparing the forecasted values against a threshold \(\delta\), calibrated to label a fixed proportion \(r\), as described in Equation (\ref{eq10}), of the validation dataset. The anomaly proportions are set as follows: 
\begin{itemize}
    \item \(r = 0.5\) for SMD, 
    \item \(r = 1\) for MSL and PSM,
    \item \(r = 4\) for SWat and SMAP.
\end{itemize}

\noindent
\paragraph{\textbf{Hyperparameter Configuration}.} The model's performance is governed by three hyperparameters: \(\lambda\), \(\alpha\), and \(\beta\).
\begin{itemize}
    \item \(\lambda\) balances the components of the loss function and is fixed at \(10^{-3}\) for all datasets.
    % \item \(\alpha\): Determines the top percentage of dominant frequency spectrums.
    % \item \(\beta\): Regulates the contribution of the time-invariant component.
    \item \(\alpha\) determines the top percentage of dominant frequency spectrums. The best-performing values of \(\alpha\) are dataset-specific: \( 0 \) for the PSM dataset, \( 0.1 \) for the SMD and SWaT datasets, and \( 0.5 \) for the SMAP and SMAP datasets. These values are summarized in Table~\ref{table:ablationAlpha}.
    
    \item \(\beta\) regulates the contribution of the time-invariant component. The optimal values for \(\beta\) are \( 0.5 \) for PSM, \( 0.8 \) for SWaT, \( 0.1 \) for SMD,\( 0 \) for MSL, and \( 0.3 \) for SMAP. These values, presented in Table~\ref{table:ablationBeta}.

\end{itemize}

\begin{table}[h!]
\centering
\caption{GRU layer counts for variant and invariant encoders.}
\footnotesize 
\begin{tabular}{l|c|c|c|c|c}
\toprule
\hline
\textbf{Dataset}         & \textbf{PSM} & \textbf{MSL} & \textbf{SMAP} & \textbf{SWaT} & \textbf{SMD} \\ \hline
Variant Encoder (Layers) & 4            & 12           & 8             & 14            & 6            \\ 
Invariant Encoder (Layers) & 2          & 8            & 2             & 8             & 2            \\ \hline
\bottomrule
\end{tabular}

\label{tab:gru_layers}
\end{table}
% \textcolor{red}{Add the valeur of $\alpha$ and $\beta$ for each datasets}
\noindent
\paragraph{\textbf{Training Setup}.} The ADAM optimizer \cite{kingma2015adam} is employed for training with an initial learning rate of \(10^{-2}\). The training process utilizes a batch size of 128 and incorporates early stopping after 10 epochs. All experiments are implemented using PyTorch \cite{paszke2019pytorch} and executed on an NVIDIA A6000 GPU.

\noindent
\paragraph{\textbf{Model Architecture}.} The model architecture of KoopAGRU comprises two specialized encoders:
\begin{enumerate}
    \item \textbf{Variant Encoder}, which extracts dataset-specific features using linear layers (100 and 128 units), ReLU activations, GRU layers with 128 hidden units, and dropout (rate: 0.01). The number of GRU layers is dataset-specific, as summarized in Table~\ref{tab:gru_layers}.
    \item \textbf{Invariant Encoder}, which captures shared patterns across datasets through linear layers, ReLU activations, GRU layers (128 hidden units), and a final linear layer for dimension matching. The GRU layer count is also dataset-specific, as shown in Table~\ref{tab:gru_layers}.
\end{enumerate}
The number of GRU layers in both encoders is dataset-specific, as summarized in Table~\ref{tab:gru_layers}.

\noindent
\paragraph{\textbf{Metrics}.} 
In anomaly detection, precision, recall, and F1-score are crucial evaluation metrics due to the inherent imbalance in datasets, where anomalies are rare compared to normal instances. These metrics are defined as follows:

\begin{equation}
\text{Precision} = \frac{\text{TP}}{\text{TP} + \text{FP}}
\end{equation}

\begin{equation}
\text{Recall} = \frac{\text{TP}}{\text{TP} + \text{FN}}
\end{equation}

\begin{equation}
\text{F1-Score} = 2 \cdot \frac{\text{Precision} \cdot \text{Recall}}{\text{Precision} + \text{Recall}}
\end{equation}

Here, \textbf{TP} (True Positives) represents the number of correctly identified anomalies, \textbf{FP} (False Positives) represents the normal points incorrectly classified as anomalies, and \textbf{FN} (False Negatives) represents the anomalies that the model failed to detect. Precision reflects the accuracy of anomaly predictions, recall indicates the model's ability to identify all actual anomalies, and the F1-score balances the trade-off between these two metrics. These measures are particularly suitable for anomaly detection tasks due to the critical need to minimize both false positives and false negatives in imbalanced datasets.

\subsection{Baselines} We extensively compare our model, KoopAGRU, with 20 baselines, including:
LSTM \cite{hochreiter1997lstm}
Transformer \cite{vaswani2017transformer},
LogTrans \cite{li2019logtrans},
TCN \cite{bai2018tcn},
Reformer \cite{kitaev2020reformer},
Informer \cite{zhou2021informer},
Anomaly \cite{xu2022anomalytransformertimeseries},
Pyraformer \cite{liu2021pyraformer},
Autoformer \cite{wu2021autoformer},
LSSL \cite{liu2022lssl},
Stationary \cite{zhou2022stationary},
DLinear \cite{zhang2023dlinear},
ETSformer \cite{woo2023etsformer},
LightTS \cite{zhou2023lightts},
FEDformer \cite{zhou2023fedformer},
TimesNet \cite{wu2023timesnet},
CrossFormer \cite{wang2023crossformer},
PatchTST \cite{nie2023patchtst}, and
ModernTCN \cite{donghao2024moderntcn}.

\begin{table}[h!]
\centering
\caption{Top 5 Average F1-scores (\%)  across all 5 datasets.} 
\resizebox{\columnwidth}{!}{%
\begin{tabular}{c|c|c|c|c|c}
\toprule
\hline
Metric                 & KoopAGRU& ModernTCN  & TimesNet & TimesNet& FEDformer \\
                       & (Ours) &    &  (R) &  (I) &  \\
\hline
Avg F1-score (\%)            & \textbf{\textcolor{red}{90.88}} & \textcolor{blue}{86.62}            & 86.34               & 85.49               & 84.97                  \\
\hline
\bottomrule
\end{tabular}
}
\label{tab:top6avgf1}
\end{table}

\begin{table*}[htbp]
% \begin{table}[htbp]
\centering
\caption{Comparison of models on various datasets. The best values are highlighted in red, while the second-best values are shown in blue.$^*$ For a fair comparison, the joint criterion in the Anomaly Transformer has been replaced with the reconstruction error.}
\resizebox{\textwidth}{!}{%
    \begin{tabular}{l|ccc|ccc|ccc|ccc|ccc|c}
    \toprule
        \hline
        \textbf{Datasets} & \multicolumn{3}{c|}{\textbf{SMD}} & \multicolumn{3}{c|}{\textbf{MSL}} & \multicolumn{3}{c|}{\textbf{SMAP}} & \multicolumn{3}{c|}{\textbf{SWaT}} & \multicolumn{3}{c|}{\textbf{PSM}} & \textbf{Avg F1} \\
        \hline
        \textbf{Metrics}
& \textbf{Prec.} & \textbf{Rec.} & \textbf{F1} & \textbf{Prec.} & \textbf{Rec.} & \textbf{F1} & \textbf{Prec.} & \textbf{Rec.} & \textbf{F1} & \textbf{Prec.} & \textbf{Rec.} & \textbf{F1} & \textbf{Prec.} & \textbf{Rec.} & \textbf{F1} & \textbf{\%}
        \\
% \begin{tabular}{|c|ccc|ccc|ccc|ccc|ccc|c|}
% \hline
% \textbf{Datasets} & \multicolumn{3}{c|}{\textbf{SMD}} & \multicolumn{3}{c|}{\textbf{MSL}} & \multicolumn{3}{c|}{\textbf{SMAP}} & \multicolumn{3}{c|}{\textbf{SWaT}} & \multicolumn{3}{c|}{\textbf{PSM}} & \textbf{Avg F1} \\
% \textbf{Metrics} & \textbf{P} & \textbf{R} & \textbf{F1} & \textbf{P} & \textbf{R} & \textbf{F1} & \textbf{P} & \textbf{R} & \textbf{F1} & \textbf{P} & \textbf{R} & \textbf{F1} & \textbf{P} & \textbf{R} & \textbf{F1} & \textbf{(\%)} \\
\hline
LSTM (1997) & 78.52 & 65.47 & 71.41 & 78.04 & 86.22 & 81.93 & 91.06 & 57.49 & 70.48 & 78.06 & 91.72 & 84.34 & 69.24 & 99.53 & 81.67 & 77.97 \\
Transformer (2017) & 83.58 & 76.13 & 79.56 & 71.57 & 87.37 & 78.68 & 89.37 & 57.12 & 69.70 & 68.84 & 96.53 & 80.37 & 62.75 & 96.56 & 76.07 & 76.88 \\
LogTrans (2019) & 83.46 & 70.13 & 76.21 & 73.05 & 87.37 & 79.57 & 89.15 & 57.59 & 69.97 & 68.67 & 97.32 & 80.52 & 63.06 & 98.00 & 76.74 & 76.60 \\
TCN (2019) & 84.06 & 79.07 & 81.49 & 75.11 & 82.44 & 78.60 & 86.90 & 59.23 & 70.45 & 76.59 & 95.71 & 85.09 & 54.59 & 99.77 & 70.57 & 77.24 \\
Reformer (2020) & 82.58 & 69.24 & 75.32 & 85.51 & 83.31 & 84.40 & 90.91 & 57.44 & 70.40 & 72.50 & 96.53 & 82.80 & 59.93 & 95.38 & 73.61 & 77.31 \\
Informer (2021) & 86.60 & 77.23 & 81.65 & 81.77 & 86.48 & 84.06 & 90.11 & 57.13 & 69.92 & 70.29 & 96.75 & 81.43 & 64.27 & 96.33 & 77.10 & 78.83 \\
Anomaly* (2021) & 88.91 & 82.23 & 85.49 & 79.61 & 87.37 & 83.31 & 81.95 & 58.11 & 71.18 & 72.51 & 97.32 & 83.10 & 68.35 & 94.72 & 79.40 & 80.50 \\
Pyraformer (2021) & 85.61 & 80.61 & 83.04 & 83.81 & 85.93 & 84.86 & 92.54 & 57.71 & 71.09 & 87.92 & 96.00 & 91.78 & 71.67 & 96.02 & 82.08 & 82.57 \\
Autoformer (2021) & 88.06 & 82.35 & 85.11 & 77.27 & 80.92 & 79.05 & 90.40 & 58.62 & 71.12 & 89.85 & 95.81 & 92.74 & 99.08 & 88.15 & 93.29 & 84.26 \\
LSSL (2022) & 78.51 & 65.32 & 71.31 & 77.55 & 88.18 & 82.53 & 89.43 & 53.43 & 66.90 & 79.05 & 93.72 & 85.76 & 66.02 & 92.93 & 77.20 & 76.74 \\
Stationary (2022) & 88.33 & 81.21 & 84.62 & 68.55 & 89.14 & 77.50 & 89.37 & 59.02 & 71.09 & 68.03 & 96.75 & 79.88 & 97.82 & 96.76 & \textbf{\textcolor{blue}{97.29}} & 82.08 \\
DLinear (2023) & 83.62 & 71.52 & 77.10 & 84.34 & 85.42 & 84.88 & 92.32 & 55.41 & 69.26 & 80.91 & 95.30 & 87.52 & 98.28 & 89.26 & 93.55 & 82.46 \\
ETSformer (2023) & 87.44 & 79.23 & 83.13 & 85.13 & 84.93 & 85.04 & 92.25 & 55.75 & 69.50 & 90.02 & 80.36 & 84.91 & 99.31 & 85.28 & 91.76 & 82.87 \\
LightTS (2023) & 87.10 & 78.42 & 82.53 & 82.40 & 75.78 & 78.95 & 92.58 & 55.27 & 69.21 & 91.18 & 94.72 & \textbf{\textcolor{blue}{93.33}} & 98.37 & 95.97 & 97.15 & 84.23 \\
FEDformer (2023) & 87.95 & 82.39 & 85.08 & 77.14 & 80.07 & 78.57 & 90.47 & 58.10 & 70.76 & 90.17 & 96.42 & 93.19 & 97.31 & 97.16 & 97.23 & 84.97 \\
TimesNet (I) (2023) & 87.76 & 82.63 & 85.12 & 82.97 & 85.42 & 84.18 & 91.50 & 57.80 & 70.85 & 88.31 & 96.24 & 92.10 & 98.22 & 92.21 & 95.21 & 85.49 \\
TimesNet (R) (2023) & 88.66 & 83.14 & 85.81 & 83.92 & 86.42 & \textbf{\textcolor{blue}{85.15}} & 92.52 & 58.29 & \textbf{\textcolor{blue}{71.52}} & 86.76 & 97.32 & 91.74 & 98.19 & 96.76 & \textbf{\textcolor{red}{97.47}} & 86.34 \\
CrossFormer (2023) & 83.60 & 76.61 & 79.70 & 84.68 & 83.71 & 84.19 & 92.04 & 55.37 & 69.14 & 88.49 & 93.48 & 90.92 & 97.16 & 89.73 & 93.30 & 83.45 \\
PatchTST (2023) & 87.42 & 81.65 & 84.44 & 84.07 & 86.23 & 85.14 & 92.43 & 57.51 & 70.91 & 80.70 & 94.93 & 87.24 & 98.87 & 93.99 & 96.37 & 84.82 \\
ModernTCN (2024) & 87.46 & 83.85 & \textbf{\textcolor{blue}{85.81}} & 83.94 & 85.93 & 84.92 & 93.17 & 57.69 & 71.26 & 91.83 & 95.98 & 93.86 & 98.09 & 96.38 & 97.23 & \textbf{\textcolor{blue}{86.62}} \\
% Chimera (2024) & 87.74 & 83.29 & 85.46 & 84.01 & 86.83 & \textbf{\textcolor{red}{85.39}} & 93.05 & 58.12 & \textbf{\textcolor{blue}{71.55}} & 92.18 & 95.93 & \textbf{\textcolor{red}{94.01}} & 97.30 & 96.19 & 96.74 & \textbf{\textcolor{blue}{86.69}} \\    
\hline
% KopaFF (Ours) & 88.78 & 86.44 & \textbf{\textcolor{red}{87.59}} & 91.51 & 79.67 & \textbf{\textcolor{blue}{85.18}} & 77.49 & 88.07 & \textbf{\textcolor{red}{82.45}} & 96.74 & 92.65 & \textcolor{red}{94.65} & 95.44 & 97.85 & 96.63 & \textbf{\textcolor{red}{89.02}} \\
        % KoopAGRU (Ours) & 88.78 & 86.44 & \textbf{\textcolor{red}{87.59}} & 91.51 & 79.67 & \textbf{\textcolor{blue}{85.18}} & 77.49 & 88.07 & \textbf{\textcolor{red}{82.45}} & 96.74 & 92.65 & \textcolor{red}{94.65} & 98.15 & 95.47 & 96.79 & \textbf{\textcolor{red}{89.33}} \\
        KoopAGRU (Ours) & 89.60 & 90.64 & \textbf{\textcolor{red}{90.12}} & 91.05 & 80.04 & \textbf{\textcolor{red}{85.19}} & 79.28 & 96.59 & \textbf{\textcolor{red}{87.09}} & 96.96 & 92.79 & \textbf{\textcolor{red}{94.83}} & 98.53 & 95.84 & 97.16 & \textbf{\textcolor{red}{90.88}} \\

\hline
\bottomrule
\end{tabular}
}
\label{tab:results}
\end{table*}
\subsection{Discussion of Results}
\paragraph{\textbf{Overall results}.} Table \ref{tab:results} provides a detailed comparison of our proposed model KoopAGRU, with state-of-the-art methods across multiple benchmark datasets for anomaly detection. Achieving an average F1-score of \textbf{90.88\%}, KoopAGRU sets a new benchmark, outperforming all other models considered. This improvement is particularly notable over advanced Transformer-based models such as \textbf{FEDformer (84.97\%)} and \textbf{Autoformer (84.26\%)}, as well as recent innovations like \textbf{TimesNet (86.34\%)} and \textbf{ModernTCN (86.62\%)}. The high performance of KoopAGRU underscores the effectiveness of integrating Fast Fourier Transform (FFT), Deep Dynamic Mode Decomposition (DeepDMD), and Koopman theory, along with GRU encoders and the new hyperparameter \(\beta\), which collectively enhance the model's ability to detect anomalies in complex, nonlinear temporal data.
\\
\paragraph{\textbf{SWaT, SMD, MSL and SMAP datasets}.} KoopAGRU demonstrates especially high performance on the \textbf{SWaR}, \textbf{SMD}, \textbf{MSL} and \textbf{SMAP} datasets, achieving F1-scores of \textbf{94.83}, \textbf{90.12}, \textbf{85.19\%} and \textbf{87.09\%}, respectively. These datasets are characterized by anomalies that manifest as disruptions to regular periodic patterns, which aligns with KoopAGRU’s design. The model’s architecture leverages FFT to decompose time-series data into time-variant and time-invariant components, enabling it to capture these periodic behaviors effectively. By employing GRU encoders, KoopAGRU learns Koopman observables across multiple temporal scales, enhancing its capability to detect anomalies that disrupt underlying periodic structures.
\\
\paragraph{\textbf{PSM dataset}.} On the \textbf{PSM} dataset, KoopAGRU achieves an F1-score of \textbf{97.03\%}, indicating a modest improvement compared to other state-of-the-art models. This incremental enhancement is likely influenced by the dataset's inherent characteristics. Anomalies in the PSM dataset are predominantly irregular and sporadic, often appearing as sudden spikes, drops, or external disturbances. While KoopAGRU is highly effective in modeling periodic and frequency-based patterns, its emphasis on periodicity analysis may reduce its sensitivity to these non-periodic and unpredictable anomalies, resulting in relatively smaller performance gains.

\subsection{Ablation Study}
This section presents an ablation study to systematically evaluate the influence of key parameters of KoopAGRU on performance and resource efficiency. The analysis focuses on the hyperparameters \( \alpha \), \( \beta \), and \( \lambda \), which control optimization dynamics, parameter balancing, and regularization, respectively. Furthermore, the study examines the impact of the anomaly ratio \( r \), which defines the proportion of data classified as anomalies. By considering both performance metrics and computational efficiency, this investigation provides a comprehensive understanding of how these parameters contribute to KoopAGRU's sensitivity, robustness, and overall effectiveness in anomaly detection.

\subsubsection{Hyperparameter \(\alpha\)}
The parameter $\alpha$ is used to select the top percent of dominant frequency spectrums. Fourier analysis, as defined in Equation~(\ref{eq1}) is employed to identify globally shared and localized frequency spectrums across different periods, separating the components of the series. 
As illustrated in Table \ref{table:ablationAlpha}, the optimal $\beta$ values for each
dataset were selected, and $\alpha$ was varied to examine its effect on the F1-scores. By setting $\alpha$ to different values, we can observe how the performance metrics, particularly the F1-score, are affected by the influence of dominant frequency spectrums. 
The results show that the F1-score generally decreases when $\alpha$ is set to lower values, indicating that the dominant frequency spectrums contribute positively to the model's performance. Notably, the SMAP dataset experiences a significant drop in F1-score with $\alpha = 0$, demonstrating a strong reliance on the dominant frequency spectrums. However, the effect varies across datasets, suggesting that both the identification and selection of dominant frequency spectrums (represented by $\alpha$) play crucial roles in determining the model's effectiveness for these datasets.
\begin{table}[h]
\centering
\footnotesize 
\caption{Comparison of ablation F1-scores of KoopAGRU with different values of $\alpha$ for different datasets. The best values are in red and the second-best values are in blue.}
\begin{tabular}{c|c|c|c|c|c}
\toprule
\hline
\textbf{Alpha} & \textbf{PSM} & \textbf{SWAT} & \textbf{SMD} & \textbf{SMAP} & \textbf{MSL} \\
\textbf{($\alpha$)} & (\(\beta\) = 0.5) & (\(\beta\) = 0.8) & (\(\beta\) = 0.5) & (\(\beta\)= 0.3) & (\(\beta\) = 0) \\
\hline
0   & \textcolor{red}{97.16} & \textcolor{blue}{93.28} & 81.06 & 67.28 & 82.07 \\
0.1 & \textcolor{blue}{96.48} & \textcolor{red}{94.83} & \textcolor{blue}{81.15} & \textcolor{blue}{71.07} & \textcolor{red}{85.19} \\
0.5 & 92.87 & 83.28 & \textcolor{red}{90.12} & \textcolor{red}{87.09} & 81.48 \\

1   & 94.00 & 84.11 & 77.79 & 67.80 & \textcolor{blue}{83.44} \\
\hline
\bottomrule
\end{tabular}%

\label{table:ablationAlpha}
\end{table}

\subsubsection{Hyperparameter \(\beta\)}
The parameter $\beta$ regulates the influence of the time-invariant component within the model, as defined in Equation~(\ref{eq_beta}). As demonstrated in Table \ref{table:ablationBeta}, the optimal $\alpha$ values for each dataset were selected, and $\beta$ was varied to examine its effect on the F1-scores. The findings reveal that modifications to $\beta$ can significantly alter performance. For instance, in the PSM dataset (where $\alpha = 0$), a slight increase in $\beta$ from 0 to 0.8 results in a minor enhancement in the F1-score, highlighting the sensitivity of this dataset to the time-invariant component. On the other hand, the SWAT dataset (with $\alpha = 0.1$) exhibits a substantial decline in the F1-score at $\beta = 0$, indicating the crucial role of the time-invariant component for its performance. Notably, the SMAP dataset (where $\alpha = 0.5$) achieves the highest F1-score at $\beta = 0.3$, underscoring the necessity of fine-tuning $\beta$ to achieve optimal performance. The differing impact of $\beta$ across these datasets underscores the importance of meticulously calibrating this parameter to enhance the model's efficacy.

\begin{table}[h]
\centering
\footnotesize 
\caption{Comparison of ablation F1-scores with different values of $\beta$ for different datasets. The best values are in red and the second-best values are in blue.}
\begin{tabular}{c|c|c|c|c|c}
\toprule
\hline
\textbf{Beta} & \textbf{PSM} & \textbf{SWAT} & \textbf{SMD} & \textbf{SMAP} & \textbf{MSL} \\
\textbf{($\beta$)} & \textbf{($\alpha=0$)} & \textbf{($\alpha=0.1$)} & \textbf{($\alpha=0.5$)} & \textbf{($\alpha=0.5$)} & \textbf{($\alpha=0.1$)} \\
\hline
0   & 96.75 & 83.60 & 84.10 & 67.28 & \textcolor{red}{85.19} \\
0.1 & 96.71 & 84.33 & \textcolor{red}{90.12} & 72.51 & 81.31 \\
0.3 & 96.69 & 84.17 & 88.15 & \textcolor{red}{87.09} & 83.26 \\
0.5 & \textcolor{red}{97.16} & \textcolor{blue}{89.61} & 83.05 & 67.49 & 81.50 \\
0.8 & \textcolor{blue}{97.03} & \textcolor{red}{94.83} & 85.68 & \textcolor{blue}{80.73} & 83.31 \\

1   & 96.72 & 84.23 & \textcolor{blue}{87.19} & 67.55 & \textcolor{blue}{85.18} \\
\hline
\bottomrule
\end{tabular}

\label{table:ablationBeta}
\end{table}
%%%%%%%%%%%%%%%%%%%%%%%%%%%%%%%%%%%%%%%%%%%%%%%%%%%
\subsubsection{Hyperparameter \(\lambda\)}
In this ablation study, we examine the impact of the regularization coefficient \( \lambda \) on model performance, as defined in Equation~(\ref{eq_lambda}), with the hyperparameters \( \alpha \) and \( \beta \) fixed at their optimal values. Table \ref{table:lambdaFScoreComparison} demonstrates that varying \( \lambda \) significantly influences F1-scores across datasets such as PSM, SWaT, SMD, SMAP, and MSL. The results indicate that \( \lambda = 10^{-3} \) consistently strikes the ideal balance between model complexity and accuracy. Higher \( \lambda \) values cause underfitting, while lower values lead to overfitting. This highlights the importance of fine-tuning \( \lambda \) to ensure that the Koopman operator matrices are both effective and generalizable, while \( \alpha \) and \( \beta \) remain optimized.

\begin{table}[ht]
\centering
\caption{Comparison of F1-scores based on different \(\lambda\) values.  The best values are in red and the second-best values are in blue.}
\footnotesize
\resizebox{\columnwidth}{!}{%
\begin{tabular}{c|c|c|c|c|c}
\toprule
\hline
\textbf{Lambda} & \textbf{PSM} & \textbf{SWaT} & \textbf{SMD} & \textbf{SMAP} & \textbf{MSL} \\
  (\textbf{$\lambda$})& \textbf{($\alpha = 0.1$,} & \textbf{($\alpha = 0.5$,} & \textbf{($\alpha = 0.1$,} & \textbf{($\alpha = 0.5$,} & \textbf{($\alpha = 0.1$,} \\

 & \textbf{$\beta = 0.8$)} & \textbf{$\beta = 0.8$)} & \textbf{$\beta = 0.1$)} & \textbf{$\beta = 0.3$)} & \textbf{$\beta = 0$)} \\

\hline
\textbf{$10^{-5}$} & 96.71 & 93.27 & 78.00 & 67.66 & 81.26 \\
\textbf{$10^{-4}$} & 96.68 & 82.77 & \textcolor{blue}{88.89} & 85.13 & 81.78 \\

\textbf{$10^{-3}$} & \textcolor{red}{97.16} & \textcolor{red}{94.83} & \textcolor{red}{90.12} & \textcolor{red}{87.09} & \textcolor{red}{85.19} \\

\textbf{$10^{-2}$} & 96.68 & 84.11 & 76.70 & 79.85 & \textcolor{blue}{82.76} \\
\textbf{$10^{-1}$} & \textcolor{blue}{96.71} & 85.03 & 79.36 & 67.26 & 81.79 \\
\hline
\bottomrule
\end{tabular}%
}

\label{table:lambdaFScoreComparison}
\end{table}

\begin{table}[h!]
\centering
\caption{F1-scores for different normalisation and denormalisation configurations across datasets, split by Time-Variant and Time-Invariant components.  The best values are in red and second-best values are in blue.}
\footnotesize  % Set font size to 9-point
\resizebox{\columnwidth}{!}{%
\begin{tabular}{c|c|c|c|c|c|c|c|c}
\toprule
\hline
\multicolumn{2}{c|}{\textbf{Time-var}} & \multicolumn{2}{c|}{\textbf{Time-inv}} & \multicolumn{5}{c}{\textbf{Dataset}} \\ \hline
\textbf{Nor.} & \textbf{Den.} & \textbf{Nor.} & \textbf{Den.} & \textbf{PSM} & \textbf{SWaT} & \textbf{SMD} & \textbf{SMAP} & \textbf{MSL} \\ \hline
 No & No & No & No & 96.09      & 87.62     & 78.45      & 67.17       & 81.75     \\ 
 Yes & No & Yes & No & 90.64      &  \textcolor{blue}{92.94}   & 88.94      & 67.83 & 82.04 \\ 
 Yes & Yes & Yes & Yes & \textcolor{blue}{96.35}      & 89.51  & \textcolor{blue}{89.27}       & \textcolor{blue}{68.12} & \textcolor{blue}{83.27}\\
 Yes & Yes & Yes & No & \textcolor{red}{97.16}      & \textcolor{red}{94.83}       & \textcolor{red}{90.12}      & \textcolor{red}{87.09}       & \textcolor{red}{85.19}      \\ \hline
 \bottomrule
\end{tabular}
}

\label{tab:norm_denorm}
\end{table}

%%%%%%%%%%%%%%%%%%%%%%%%%%%%%%%%%%%%%%%%%%%%%%%%%%%
\subsubsection{Anomaly Ratio \(r\)}
The optimal $\alpha$ and $\beta$ values for each dataset were selected, and the anomaly ratio was varied to examine its impact on F1-scores, as defined in Equation~(\ref{eq10}). Table \ref{table:abilationThreshold} illustrates how adjusting the anomaly ratio influences performance metrics. The results demonstrate that the anomaly ratio significantly affects F1-scores across different datasets. For example, in the PSM dataset (with $\alpha = 0$ and $\beta = 0.5$), the F1-score remains relatively stable, with the highest score observed at an anomaly ratio of 1. Conversely, the SWaT dataset (with $\alpha = 0.1$ and $\beta = 0.8$) achieves its highest F1-score at an anomaly ratio of 4, indicating that higher anomaly ratios enhance performance for this dataset. The SMAP dataset (where $\alpha = 0.5$ and $\beta = 0.3$) records its highest F1-score at an anomaly ratio of 4, highlighting the importance of tuning the anomaly ratio for optimal performance. The differing impact of the anomaly ratio across these datasets underscores the necessity of calibrating this parameter to improve the model's effectiveness. The results indicate that different datasets respond variably to changes in the anomaly ratio, reflecting the diverse nature of the datasets and the importance of selecting suitable parameters for effective anomaly detection.

\begin{table}[ht]
\centering
\caption{Comparison of F1-scores based on different anomaly ratios. The best values are in red and the second-best values are in blue.}
\resizebox{\columnwidth}{!}{%
\begin{tabular}{c|c|c|c|c|c}
\toprule
\hline
\textbf{Anomaly Ratio} & \textbf{PSM} & \textbf{SWaT} & \textbf{SMD} & \textbf{SMAP} & \textbf{MSL} \\
  (\textbf{$r$})& \textbf{($\alpha = 0$,} & \textbf{($\alpha = 0.1$,} & \textbf{($\alpha = 0.5$,} & \textbf{($\alpha = 0.5$,} & \textbf{($\alpha = 0.1$,} \\

 & \textbf{$\beta = 0.5$)} & \textbf{$\beta = 0.8$)} & \textbf{$\beta = 0.1$)} & \textbf{$\beta = 0.3$)} & \textbf{$\beta = 0$)} \\
\hline
0.5 & 96.48 & \textcolor{blue}{90.61} & \textcolor{red}{90.12} & 67.37 & \textcolor{blue}{82.08} \\
1 & \textcolor{red}{97.16} & \textcolor{blue}{90.61} & \textcolor{blue}{85.31} & 67.20 & \textcolor{red}{85.18} \\
4 & \textcolor{blue}{96.55} & \textcolor{red}{94.83} & 63.91 & \textcolor{red}{87.09} & 80.66 \\
5 & 96.08 & 90.00 & 59.26 & \textcolor{blue}{84.65} & 78.25 \\
\hline
\bottomrule
\end{tabular}%
}
\label{table:abilationThreshold}
\end{table}

\begin{table*}[ht]
\centering
\caption{Model performance comparison across different datasets}
\footnotesize
\resizebox{\textwidth}{!}{%
\begin{tabular}{l|l|r|r|r|r|r}
\toprule
\hline
\textbf{Dataset} & \textbf{Model} & \makecell{\textbf{Number of} \\ \textbf{Parameters}} & \makecell{\textbf{Training Time} \\ \textbf{(seconds)}} & \makecell{\textbf{RAM Usage} \\ \textbf{(MB)}} & \makecell{\textbf{GPU Usage} \\ \textbf{(MB)}} & \makecell{\textbf{Testing Time} \\ \textbf{(seconds)}} \\
\hline
\multirow{4}{*}{SWaT} & Autoformer & 323,379 & 862.30 & 17,425.98 & 17.48 & 183.05 \\
 & TimesNet & 4,700,851 & 1477.07 & 19,116.41 & 35.72 & 279.45 \\
 & Transformer & 325,683 & 543.31 & 15,690.57 & 17.49 & 97.75 \\
 & KoopAGRU (Ours) & 1,605,402 & 3039.47 & 5,197.27 & 17.77 & 153.13 \\
\hline
\multirow{4}{*}{SMD} & Autoformer & 316,710 & 53.59 & 829.24 & 17.46 & 5.01 \\
 & TimesNet & 4,697,510 & 124.01 & 1,074.05 & 38.52 & 7.67 \\
 & Transformer & 319,014 & 36.37 & 690.68 & 17.47 & 3.67 \\
 & KoopAGRU (Ours) & 694,032 & 60.75 & 18.56 & 7.78 & 4.48 \\
\hline
\multirow{4}{*}{SMAP} & Autoformer & 310,041 & 363.90 & 7,991.14 & 17.43 & 112.87 \\
 & TimesNet & 28,133,145 & 1753.46 & 9,230.68 & 127.62 & 329.39 \\
 & Transformer & 312,345 & 195.38 & 7,688.74 & 17.44 & 59.71 \\
 & KoopAGRU (Ours) & 874,262 & 1137.86 & 13.37 & 10.25 & 72.81 \\
\hline
\multirow{4}{*}{PSM} & Autoformer & 310,041 & 207.45 & 2,289.70 & 17.43 & 39.90 \\
 & TimesNet & 4,694,169 & 374.68 & 2,603.40 & 37.47 & 65.60 \\
 & Transformer & 312,345 & 117.58 & 2,071.07 & 17.44 & 20.91 \\
 & KoopAGRU (Ours) & 1,293,950 & 502.50 & 18.45 & 15.39 & 30.57 \\
\hline
\multirow{4}{*}{MSL} & Autoformer & 325,431 & 418.37 & 3,771.96 & 17.49 & 26.67 \\
 & TimesNet & 75,223 & 28.67 & 3,301.14 & 16.54 & 15.43 \\
 & Transformer & 327,735 & 279.41 & 3,606.20 & 17.50 & 14.52 \\
 & KoopAGRU (Ours) & 1,430,930 & 397.81 & 2.33 & 16.04 & 19.31 \\
\hline
\bottomrule
\end{tabular}
}

\label{tab:model_performance}
\end{table*}

\subsubsection{Resource Efficiency}
As shown in Table \ref{tab:model_performance}, a comparative analysis of four key performance metrics, namely the number of parameters, testing time in seconds, GPU usage in megabytes, and RAM usage in megabytes, indicates several advantageous features of our proposed model compared to existing alternatives. KoopAGRU's number of parameters is relatively low compared to the existing models, which indicates enhanced memory efficiency and reduced computational.
Low resource requirements are important for resource-constrained environments. KoopAGRU has a considerably shorter test-time period compared to existing models implying that it can be extensively used for processing inference tasks at an accelerated rate required for real-time applications including anomaly detection.
Similarly, this analysis shows that our model moderately consumes GPU memory thus supporting its capability to accomplish difficult calculations devoid of overloading hardware. It is a significant feature when shared GPU resources become large-scale deployments. Additionally, the RAM usage profile of our model is considerably lower, hence making it possible to run on systems with limited RAM availability.
\\
\\
In conclusion, KoopAGRU has shown outstanding parameter efficiency, lower test time, and optimal use of available resources which makes it a good competitor for an efficient and scalable machine-learning solution that can be used in any practical deployment context.

\subsection{Model limitations}
Although KoopAGRU exhibits state-of-the-art performance, the model may have the following limitations:
\begin{itemize}
\item Training Time: Due to the incorporation of the projection step and the newly introduced loss function, the training time of KoopAGRU is significantly longer compared to other models. This extended training period is a result of the model's complexity and the additional computational overhead, which might limit its practicality in time-sensitive applications.
\item Limited Dataset Evaluation: Our model has been tested exclusively on five datasets, which are widely used in the literature for time series anomaly detection. While these datasets provide a robust benchmark, the performance of KoopAGRU on other types of time series data and tasks remains to be thoroughly explored.
\end{itemize}

\section{Conclusion}

In this paper, we introduced KoopAGRU, an improved deep model for time series anomaly detection that utilizes the strength of DeepDMD and Koopman theory. KoopAGRU’s ability to dynamically adapt to varying datasets enhances both its scalability and robustness through the introduction of a novel hyperparameter \(\beta\), which allows for precise tuning across different time series, while the direct construction of observables simplifies the model and reduces computational complexity. Moreover, the fixed-size Koopman operator optimizes the model size without compromising performance. Experimental results confirm that KoopAGRU achieves good results, consistently outperforming existing methods and demonstrating its efficacy as a robust and efficient solution for real-world anomaly detection tasks.

%%%%%%%%%%%%%%%%%%%%%%%%%%%%%%%  References %%%%%%%%%%%%%%%%%%%%%%%%%%%%%%%%%%%%%%
\bibliography{sample}
\bibliographystyle{IEEEtran}

%%%%%%%%%%%%%%%%%%%%%%%%%%%%%%%  Appendix %%%%%%%%%%%%%%%%%%%%%%%%%%%%%%%%%%%%%%
\clearpage
\appendix
\section{Preliminaries}
\label{appendix:A}

\noindent 

% Existing machine learning (ML) methods for transient prediction either require extensive system features (such as detailed generator models) or are limited to local prediction problems (such as the transient trajectories of a single generator). This makes them inapplicable to system-wide transient prediction problems under limited information (such as access to noisy measurements data only).

% In this article, we focus on data-driven learning models that rely only on streaming measurements from sensors (such as PMUs) across the network. We also consider some limited information, such as the topology of the network and the locations of the sensors.
\subsection{The Koopman operator}
\label{appendix:koopman}
In this section, we revisit the mathematical foundation of the Koopman Operator Theory (KOT). Let's consider the following discrete-time nonlinear system described by the relation: 
\begin{equation}\label{aeq1}
Z_{t+1} = \mathcal{F}(Z_t)
\end{equation} \\
Here, $Z_t$ exists within an m-dimensional space $M$ contained in $\mathbb{R}^m$, playing the role of system states. $\mathcal{F}$ is a continuously differentiable system map, denoted as $\mathcal{F}: M \rightarrow M$. For the purpose of this paper, we notice that $\mathcal{F}(\cdot)$ is unknown. Next, $\mathcal{F}(M)$ will stand for the universal space of all scalar-valued functions of the system states $z$.
\begin{definition}
(Koopman Operator) \\The Koopman operator, $\mathbb{U}: \mathcal{F}(\mathcal{M}) \rightarrow \mathcal{F}(\mathcal{M})$, associated with the dynamical system  \eqref{aeq1} is defined as an infinite-dimensional linear operator which acts on any scalar-valued observable function $\varphi \in \mathcal{F}(\mathcal{M})$ of the system \eqref{aeq1} as follows:
\begin{equation}\label{aeq2}
[\mathbb{U} \varphi](\boldsymbol{z})=\varphi(\mathcal{F}(\boldsymbol{z})) 
\end{equation} \\
In other terms, we define the Koopman operator, $\mathbb{U}$, as an infinite-dimensional linear operator that pairs up with the dynamical system and acts on any scalar-valued observable function $\varphi$ within $\mathcal{F}(M)$ that belongs to the system. The action of $\mathbb{U}$ on $\varphi$ is given by $[\mathbb{U}\varphi](z) = \varphi(\mathcal{F}(z))$ and the evolution of these observable functions is facilitated by $\mathbb{U}$. A noteworthy property of the operator $\mathbb{U}$ is that it maintains the linearity and invariance of the function space $\mathcal{F}(M)$.
\end{definition}
Generally speaking, the process of mapping these functions through the Koopman operator, being infinite-dimensional in nature, is referred to as "lifting". As a positive operator, $\mathbb{U}$ ensures that if $\varphi(z) \geq 0$, then $[\mathbb{U}\varphi](z)$ will also have a non-negative value. In instances where the mathematical model \eqref{aeq1} is not explicitly known, a Koopman operator can be effectively determined from data sampled from time-series experiments or sensor measurements. In actual practice, what is typically computed is an approximation of the Koopman operator that exists in finite-dimensional space.

\subsection{DeepDMD}
\label{appendix:DeepDMD}

As the $\varphi$ function of the \textit{Koopman operator} is generally difficult to compute directly due to the absence of a straightforward analytical method, DeepDMD \cite{deepdmd} has been proposed as a data-driven approach to approximate this function. By leveraging the power of neural networks, DeepDMD learns a mapping function $\psi$, which serves as an approximation of the true $\varphi$ function associated with the \textit{Koopman operator}. 
\\
The core idea behind the \textit{deepDMD} methodology is to identify this specific mapping function $\psi$. In doing so, it also seeks to uncover a corresponding linear operator—an approximate variant of the \textit{Koopman operator}, represented in equation \eqref{aeq2}. This operator fundamentally describes the evolution of observable functions within an expanded domain known as the lifted observables space.
\\
In \textit{deepDMD}, the function $\psi$ plays a crucial role as it transfers time-series data into an observable space. However, the task does not end there, as the approach also requires an inverse mapping that enables the retrieval of the original measurements from the observable space. Two approaches can accomplish this reversal. The first method involves identifying a function $\varphi$, which is essentially the reverse of $\psi$. When this $\varphi$ function is recognized, it suggests that $X_t$ can equivalently be depicted as $\varphi(\psi(X_t))$. In contrast, the reverse operation can be carried out by conjoining the actual measured states, represented by $X_t$, with observables to achieve \textit{measurement-inclusive observables}. This amalgamation results in a composite function $\Psi(X_t)$ that incorporates both $X_t$ and $\psi(X_t)$.
\begin{equation}
\Psi(X_t) = \begin{bmatrix} X_t \\ \psi(X_t) \end{bmatrix}
\end{equation} 
Interestingly, the measurements are readily available and accessible, which obviates the necessity for a complex inverse mapping when dealing with \textit{measurement-inclusive observable functions}. Therefore, we have adopted the second method for this work due to its comparative simplicity and its close resemblance to the notion of \textit{state-inclusive observables}. The learning component in this framework requires discovering a suitable observable function represented as $\Psi$, and an appropriate \textit{Koopman operator}, $\mathcal{K}$. These two should adhere to the relationship 
\begin{equation}
\Psi(X_{t+1}) = \mathcal{K}\Psi(X_t)
\end{equation} 
at each time-point $t$. Due to the application of a neural network in \textit{deepDMD}, the observable function is represented as $\Psi(X_t, \Theta)$, where $\Theta$ embodies the neural network's weights and biases.
Let's consider a series of $n+1$ measurements referred to as ‘snapshots’:
\begin{equation}
\begin{pmatrix} \mathbf{x}(1) & \mathbf{x}(2) & \dots & \mathbf{x}(n+1) \end{pmatrix}
\end{equation} 
In this scenario, we define: 
\begin{equation}
\mathbf{X}_t  \text{ as } \begin{pmatrix} \mathbf{x}(1) & \mathbf{x}(2) & \dots & \mathbf{x}(n) \end{pmatrix}
\end{equation} 
and 
\begin{equation}
\mathbf{X}_{t+1}  \text{ as } \begin{pmatrix} \mathbf{x}(2) & \mathbf{x}(3) & \dots & \mathbf{x}(n+1) \end{pmatrix}
\end{equation} 
Both $\mathbf{X}_{t}$ and $\mathbf{X}_t+$ are essentially a series of measurements taken from $N$ unique points in time. However, $\mathbf{X}_{t+1}$ is derived by moving $\mathbf{X}_t$ forward by a single time-point. The \textit{Koopman operator} relationship $\Psi(\mathbf{X}_{t+1} , \Theta) = \mathcal{K}\Psi(\mathbf{X}_t, \Theta)$ holds true only if $\mathbf{X}_{t+1}$ is an accurate shift of $\mathbf{X}_t$.
\\
The \textit{deepDMD} learning cost function can thus be defined as: 
\begin{equation}
\min_{\mathcal{K}, \Theta} \|\Psi(\mathbf{X}_{t+1}, \Theta) - \mathcal{K}\Psi(\mathbf{X}_t, \Theta)\|_F^2 + \lambda \|K\|_F^2 
\end{equation} 
where $\lambda$ ($>$ 0) represents, the weights for the regularization costs that prevent the \textit{Koopman operator} from becoming overfitted to the learned models, while also enhancing their robustness.

\vfill

\end{document}